%% file: main.tex
\documentclass[preprint, nopreprintline, 12pt]{elsarticle}

\usepackage{amssymb}
\usepackage{amsmath}
\usepackage[dvipsnames]{xcolor}
\usepackage[normalem]{ulem}
\usepackage{glossaries}
\usepackage{siunitx}
\usepackage{xspace}
\usepackage{booktabs}
\usepackage{array}
\usepackage{adjustbox}
\usepackage{multirow}
\input{glossaries}
\usepackage{pifont}
\usepackage{makecell}
\usepackage{xspace}
\usepackage{siunitx}
 \usepackage{graphicx}
 \usepackage{xcolor}
\newcommand{\etal}{\textit{et~al.}\xspace}
\newcommand{\eg}{\textit{e.g.}\xspace}

\newcommand{\Ms}{M\textsubscript{simulated}\xspace}
\newcommand{\Mr}{M\textsubscript{real}\xspace}
\newcommand{\Mf}{M\textsubscript{finetune}\xspace}
\usepackage{hyperref}
\usepackage{graphicx}
\usepackage{subcaption}
\usepackage{svg}

\usepackage{lineno} 

\journal{Medical Image Analysis}

\begin{document}

\begin{frontmatter}

\title{Eliminating Registration Bias in Synthetic CT Generation: A Physics-Based Simulation Framework}

\author[1,2]{Lukas Zimmermann} 
\author[3]{Michael Rauter}
\author[1,2]{Maximilian Schmid}
\author[1,2]{Dietmar Georg}
\author[1,2]{Barbara Knäusl}

\affiliation[1]{organization={Medical University of Vienna, Department of Radiation Oncology},
    city={Vienna},
    country={Austria}}
\affiliation[2]{organization={Christian Doppler Laboratory for Image and Knowledge Driven Precision Radiation Oncology, Medical University of Vienna},
    city={Vienna},
    country={Austria}}
\affiliation[3]{organization={University of Applied Sciences Wiener Neustadt, Competence Center for Preclinical Imaging \& Biomedical Engineering},
    city={Wiener Neustadt},
    country={Austria}}

\begin{abstract}
Supervised synthetic Computed Tomography (CT) generation from cone-beam CT (CBCT) requires registered training pairs, yet perfect registration between separately acquired scans remains unattainable. This registration bias propagates into trained models and corrupts standard evaluation metrics. This may suggest that superior benchmark performance indicates better reproduction of registration artifacts rather than anatomical fidelity. We propose physics-based CBCT simulation to provide geometrically aligned training pairs by construction, combined with evaluation using geometric alignment metrics against input CBCT rather than biased ground truth. On two independent pelvic datasets, models trained on synthetic data achieved superior geometric alignment (Normalized Mutual Information: 0.31 vs 0.22) despite lower conventional intensity scores. Intensity metrics showed inverted correlations with clinical assessment for deformably registered data, while Normalized Mutual Information consistently predicted observer preference across registration methodologies ($\rho$ = 0.31, p $<$ 0.001). Clinical observers preferred synthetic-trained outputs in 87\% of cases, demonstrating that geometric fidelity, not intensity agreement with biased ground truth, aligns with clinical requirements.
\end{abstract}




\end{frontmatter}


\input{introduction} 

\section{Materials and Methods}
\label{mat}
\input{framework}
\input{experiments}

\input{results}

\input{discussion}

\input{conclusion}

\section*{Author Contributions}
Lukas Zimmermann: Conceptualization, Data curation, Software, Formal analysis, Methodology, Writing – original draft

Michael Rauter: Software, Formal analysis, Methodology, Writing – original draft

Dietmar Georg: Writing – review and editing

Barbara Knäusl: Writing – review and editing, Funding acquisition, supervision project, administration

Maximilian Schmid: Writing – review and editing, Funding acquisition

\section*{Funding}
The financial support by the Austrian Federal Ministry of Economy, Energy and Tourism, the National Foundation for Research, Technology and Development and the Christian Doppler Research Association is gratefully acknowledged. Michael Rauter was funded as part of the RTI-strategy Lower Austria 2027.

\section*{Acknowledgments}
The authors thank the high performance computing team at the Medical University of Vienna for providing computational resources. We acknowledge the five independent observers (Data scientists = Lukas Zimmermann and Michael Rauter, Medical Physicists: Barbara Knäusl and Martin Buschmann, Radiation technologist = Kerstin Feichtinger) who participated in the visual analogue score assessment. We also thank the SynthRAD Challenge organizers for making the multi-institutional dataset available.

\section*{Declaration of generative AI and AI-assisted technologies in the manuscript preparation process}
During the preparation of this work the author(s) used claude.ai in order to improve writing and grammar. After using this tool/service, the author(s) reviewed and edited the content as needed and take(s) full responsibility for the content of the published article.

\appendix

\input{ablation_study}

\bibliography{./bibliography}

\end{document}

%% file: glossaries.tex
\newacronym{VAE}{VAE}{Variational Autoencoder}
\newacronym{gan}{GAN}{Generative Adversarial Network}
\newacronym{CBCT}{CBCT}{cone-beam computed tomography}
\newacronym{sCBCT}{sCBCT}{simulated CBCT}
\newacronym{CT}{CT}{computed tomography}
\newacronym{MRI}{MRI}{Magnetic Resonance Imaging}
\newacronym{MC}{MC}{Monte Carlo}
\newacronym{DRR}{DRR}{Digital reconstructed radiograph}
\newacronym{FOV}{FOV}{field of view}
\newacronym{RTK}{RTK}{Reconstruction Toolkit}
\newacronym[shortplural={OARs},longplural=organs at risk]{OAR}{OAR}{organ at risk}
\newacronym{phscbct}{Ph-sCBCT}{physics-driven synthetic CBCT}
\newacronym{DSC}{DSC}{Dice-Sorense coefficient}
\newacronym{MSD}{MSD}{mean surface distance}
\newacronym{HD95}{HD95}{95\% Hausdorff distance}
\newacronym{JI}{JI}{Jaccard index}
\newacronym{SPR}{SPR}{scatter-to-primary ratio}
\newacronym{MAE}{MAE}{Mean Absolute Error}
\newacronym{SSIM}{SSIM}{Structural Similarity index measure}
\newacronym{PSNR}{PSNR}{peak signal-to-noise ratio}
\newacronym{DICE}{DICE}{Dice Score}
\newacronym{H95}{H95}{Hausdorff 95\textsuperscript{th}}
\newacronym{NMI}{NMI}{normalized mutual information}
\newacronym{CC}{CC}{cross correlation}
\newacronym{IQS}{IQS}{Image Quality scale}
\newacronym{FID}{FID}{Fréchet Inception distance}
\newacronym{MMD}{MMD}{Maximum Mean Discrepancy}
\newacronym{rt}{RT}{radiotherapy}
\newacronym{sct}{sCT}{synthetic CT}
\newacronym{dir}{DIR}{deformable image registration}
\newacronym{sCT}{sCT}{synthetic CT}
\newacronym{HU}{HU}{Hounsfield unit}
\newacronym{CI95}{CI95\%}{95\% confidence interval}
\newacronym{ICC}{ICC}{intraclass correlation coefficient}

%% file: introduction.tex
\section{Introduction}
\label{intro}

Daily \gls{CBCT} imaging is the most commom used basis for image-guided adaptive \gls{rt} workflows, enabling visualization of daily anatomical changes \cite{DonaLemus2024}. However, \gls{CBCT} images suffer from scatter artifacts, respiratory and organ motion, increased noise, and poor soft tissue contrast compared to fan-beam \gls{CT} \cite{Schulze2011}. These limitations may prevent direct use of \gls{CBCT} images for critical tasks in adaptive \gls{rt} such as structure segmentation and dose calculation or introduce systematic uncertainties.\\
Deep learning approaches have emerged as a promising solution, learning the conversion from \gls{CBCT} to diagnostic-quality \gls{CT} (called \gls{sct}) using paired clinical datasets \cite{Rabe2025, DAYARATHNA2024}. This development was further accelerated by large open-source datasets, \eg established through the SynthRAD challenges~\cite{Thummerer2023, Thummerer2025}.\\
However, supervised methods trained on real acquisitions inherit the fundamental limitation that \gls{CBCT} and \gls{CT} cannot be registered without residual error \cite{Nenoff2023}. Fan-beam \glspl{CT} are acquired before treatment under optimal conditions, while \glspl{CBCT} during the treatment course capture daily anatomy with variations in organ filling, respiratory state, and patient positioning. Even sophisticated \gls{dir} algorithms cannot achieve perfect correspondence when anatomy has changed between acquisitions~\cite{Suwanraksa2023}. This registration uncertainty creates systematic bias during training. As consequence networks are trained with voxel-wise losses to suppress or blur these features to minimize spatial discrepancies, rather than learning the true intensity transformation while preserving geometric fidelity. \\
Unpaired learning methods, most notably CycleGAN~\cite{Liang2019, Liu2020}, were developed to circumvent alignment requirements using cycle-consistency losses. While this eliminates the need for paired data, it provides no explicit geometric supervision, allowing hallucination of structures or introduction of artifacts. Recent registration-aware approaches integrate deformable registration networks directly into training~\cite{Suwanraksa2023, Hu2024_SynREG}, but remain dependent on registration quality. Critically, this not only affects training but also evaluation.\\
Standard intensity metrics measure agreement with ground truth, however, when the ground truth itself contains registration errors, models that reproduce these errors are rewarded. This creates a fundamental evaluation paradox: superior benchmark performance may indicate better learning of registration artifacts rather than better anatomical fidelity. Breaking this cycle requires both bias-free training data and bias-free evaluation metrics. These considerations also raise the question of suitable evaluation metrics. Current benchmarks rely on intensity-based metrics (\gls{MAE}, \gls{PSNR}, \gls{SSIM}) \cite{SynthRAD2023}, yet for clinical applications requiring accurate image registration, such as dose accumulation, geometric alignment metrics may be at least equally important.\\
Physics-based simulation offers a different solution. Rather than correcting registration errors or avoiding paired data, \gls{sCBCT} images generated from fan-beam \gls{CT} scans are inherently aligned to their source by construction eliminating the registration problem. All image differences arise solely from modeled physical processes rather than anatomical misalignment. Proof of principle studies have explored this direction: DeepDRR demonstrated physics-based X-ray projection synthesis for 2D applications~\cite{DeepDRR2018} and SinoSynth showed that networks trained on simulated data can outperform those trained on real data~\cite{Pang2024}. However, extending these concepts to 3D volumetric \gls{CBCT} synthesis with respiratory motion modeling has not been systematically investigated.\\
This study presents a physics-based framework for generating \gls{sCBCT} images from fan-beam \gls{CT} scans, incorporating heuristic scatter modeling and respiratory motion simulation. Unlike \gls{MC} simulations, which are computationally prohibitive for large-scale data generation, our approach proved to generate \gls{sCBCT} images efficiently, enabling creation of training cohorts with guaranteed geometric correspondence. We evaluated this framework in the pelvic region on both a clinical dataset and the SynthRAD 2023 challenge dataset and assessed performance using intensity-based metrics and geometric alignment measures (\gls{NMI}, \gls{CC}). This study aims to demonstrate that models trained on simulated imaging data achieve superior geometric alignment compared to models trained on real but imperfectly registered pairs, validating that perfect anatomical correspondence during training enables preservation of spatial relationships.\\

\section{Previous work}
Recent registration-aware approaches attempted to address the registration error challenge. Suwanraksa \etal developed RegGAN for head and neck cancer patients, achieving MAE of 43.7 HU by treating imperfect alignment as "noisy labels" \cite{Suwanraksa2023}. Hu \etal proposed SynREG, a transformer-based framework validated across multiple anatomical sites \cite{Hu2024_SynREG}. However, both acknowledged that perfect alignment remains unattainable and hallucinations persist \cite{Suwanraksa2023, Hu2024_SynREG}. These findings suggest that attempting to learn and correct for registration errors simultaneously with intensity transformation remains challenging, motivating our approach of eliminating registration errors at the data generation stage.\\ 
Physics-based \gls{CBCT} simulation has been explored previously, though with different objectives and implementation strategies. Peng \etal developed SinoSynth, a physics-based degradation model operating in projection (sinogram) space that demonstrated networks trained on simulated data can outperform those trained on real data \cite{Pang2024}. However, the evaluation was performed only for the head and neck region and not for the pelvic area as well as the geometry is a simple fan-beam operating in 2D slices. They describe a random motion model in the form of rotation and resizing the volume not accuratly representing phyisological deformations. To better simulate breathing motion, which is important in the pelvic region, a physiologically plausible motion model has to be defined.\\
The DeepDRR framework \cite{DeepDRR2018, DeepDRR2019} established foundational methods for physics-based X-ray projection synthesis, demonstrating the feasibility of generating realistic 2D projections for machine learning applications. The GPU-accelerated implementation enables reduced generation times making large-scale dataset creation practical without the computational burden of Monte Carlo simulation. Our work extends these concepts to full 3D volumetric \gls{CBCT} synthesis while adding respiratory motion modeling. \\

%% file: framework.tex
\subsection{Physics simulation framework}
\label{sec:framework}
\begin{figure}[!h]
    \centering
    \includegraphics[trim={0 0 0 0}, clip, width=1.0\linewidth]{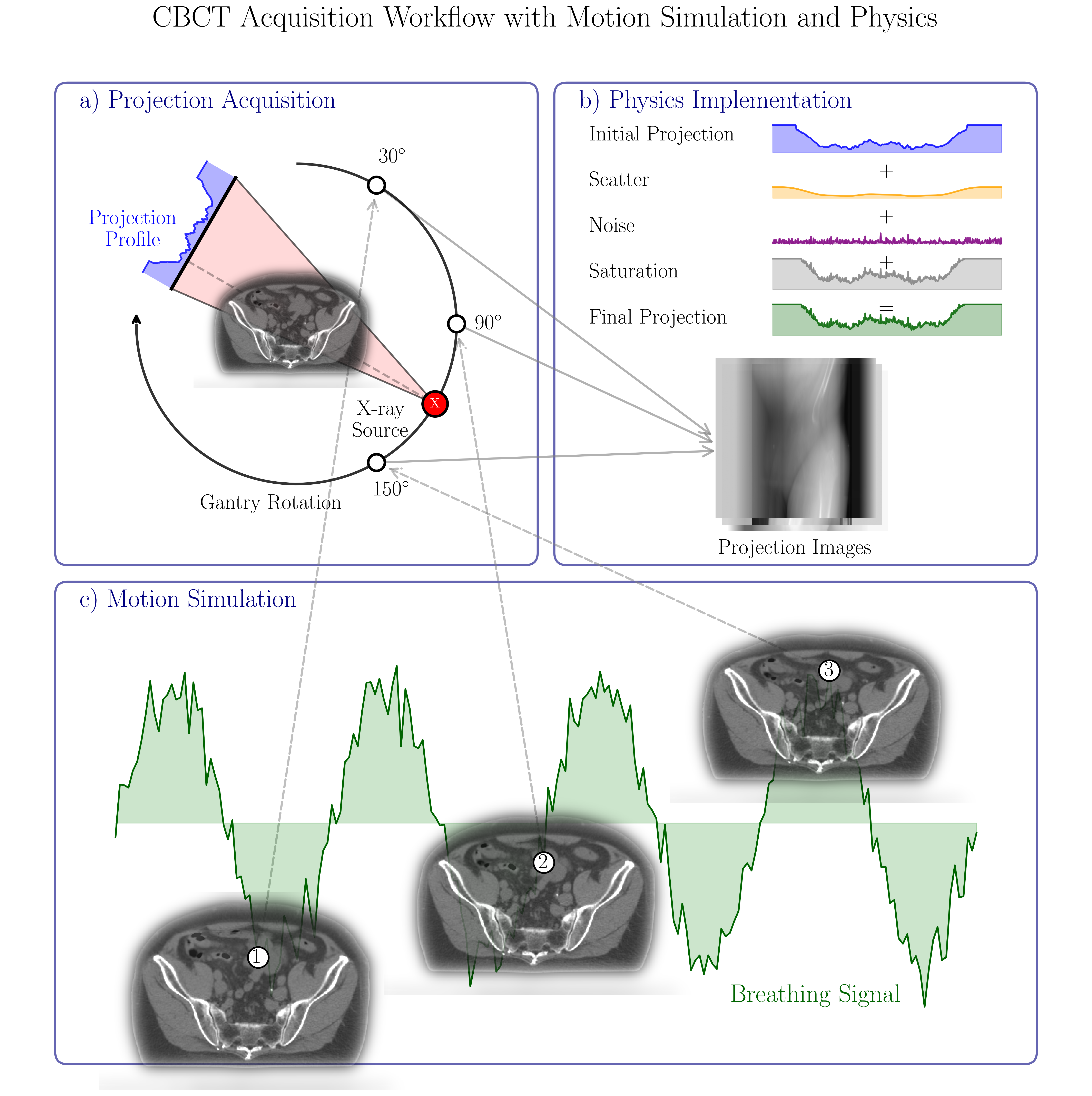}
    \caption{Simulated \gls{CBCT} generation workflow. The fan-beam \gls{CT} is deformed using a respiratory motion field to simulate breathing motion (c). Motion-adapted volumes are forward projected to create simulated X-ray projections (a), which are then reconstructed with simulated artifacts (b) to generate the final simulated \gls{CBCT}.}
    \label{fig:workflow}
\end{figure}

Simulated \gls{CBCT} images were generated through physics-based simulation of the image acquisition process for the pelvic anatomy. The approach consisted of three sequential stages: (1) respiratory motion simulation to create motion-adapted \gls{CT} volumes at different breathing phases, (2) forward projection to generate simulated X-ray projections, and (3) reconstruction with simulated artifacts such as scatter, motion, and noise (details in \ref{ch:respiratory_motion}, \ref{sec:drr_recon} and \ref{sec:recon}). An overview of the complete workflow is shown in \autoref{fig:workflow}. Scanner specific selected adjustable parameters for our experiments can be found in \autoref{tab:parameter}. General parameters were fixed and directly stated in the text. The framework can be found at \hyperlink{https://github.com/openvoxelmed/simcbctgenerator}{https://github.com/openvoxelmed/simcbctgenerator}.\\

\begin{table}[]
    \centering
    \resizebox{\linewidth}{!}{
    \begin{tabular}{ccc}
    \toprule
    & Parameter & Value \\
    \midrule
     \multirow{5}{*}{Motion Simulation} & $\tau_g$ & 200 HU/mm \\
      & bone threshold  & 200 HU \\
       &  $T_p$ & 180 ms \\
        &  $T_{hc}$ & 1.5 s \\
         &  $A_{max}$ & 5 mm \\
         \midrule
       \multirow{14}{*}{Dynamic X-ray projection}   & $\Phi$  & $4.16\cdot10^5$ photons/(mm\textsuperscript{2}$\cdot$mAs)\\
           & SPR  & 1.6 \\
           & current & 40 mA\\
           &exposure time& 40 ms\\
           &detector spacing & 0.8 mm $\times$ 0.8 mm\\
           & detector size & 409.6 mm $\times$ 409.6 mm\\
           & detector pixels & 512$\times$512\\
           & source-to-detector distance & 1536 mm\\
           & source-to-axis distance & 1000 mm\\
           & detector offset & 115 mm\\
           & start/stop angle & -180/180$^\circ$\\
           & angle increments & 0.54$^\circ$\\
& saturation factor & 2.0 \\
\midrule
    \multirow{2}{*}{CBCT reconstruction}   & volume dimensions (x$\times$y$\times$z) & $410\times410\times66$\\  
        & volume spacing (x$\times$y$\times$z) & 1 mm$\times$1 mm$\times$4 mm\\
         \bottomrule
    \end{tabular}
    }
    \caption{Selected parameters for the simulated image generation. The settings are specific for the Elekta XVI which was clinically available. }
    \label{tab:parameter}
\end{table}

\subsection{Respiratory Motion Simulation}
\label{ch:respiratory_motion}

A motion vector field was derived from the fan-beam CT individually for each patient to simulate abdominal respiratory motion. This field specified a displacement vector for each voxel position, scaled by a time-varying breathing state to generate motion-corrupted volumes representing different phases of the respiratory cycle.

\paragraph{Motion Field Derivation}
A heuristic motion field was designed to replicate characteristic abdominal breathing patterns observed during image acquisition. Our respiratory motion exhibits several key assumptions: (1) motion magnitude is strongest at the patient surface and decreases toward the body center, (2) motion direction aligns with the intensity gradient at tissue-air boundaries (perpendicular to the body surface), (3) motion is predominantly in the anterior direction with increasing attenuation for deviations from this main axis, and (4) motion is minimal in bone regions and tissue posterior to bone structures.

The motion field generation involved the following steps:

\begin{enumerate}
    \item Create a binary foreground mask identifying voxels belonging to the patient body and extract the inner boundary of the patient foreground mask to identify the body surface.
    
    \item Compute 3D gradient vectors $\nabla I(\mathbf{x})$ at boundary positions from the CT intensity field, retaining only gradients exceeding the threshold $\tau_g$ (\autoref{tab:parameter}) to create a sparse 3D vector field $\mathbf{V_s}(\mathbf{x})$ (\autoref{fig:motion_field:MF_strong_border_gradients}).
    
    \item Propagate boundary gradients throughout the volume using distance-weighted interpolation to create a dense vector field $\mathbf{V_d}(\mathbf{x})$ (\autoref{fig:motion_field:MF_propagated_gradients1}).
    
    \item Attenuate motion vectors deviating from the anterior anatomical direction using dot product weighting with the anterior direction vector, where negative dot products are set to zero. The resulting vector field $\mathbf{V_u}(\mathbf{x})$ is shown in \autoref{fig:motion_field:MF_propagated_gradients2}.
    
    \item Identify regions requiring strong motion (using a surrogate structure as reference) and regions requiring motion suppression (bone tissue identified by thresholding seen in \autoref{tab:parameter}, and tissue posterior to bone), which yields the vector field $\mathbf{V_f}(\mathbf{x})$.
    
    \item Apply region-specific attenuation to produce the final motion field $\mathbf{M}(\mathbf{x}, s_i)$, where $s_i$ is the breathing state (see \autoref{fig:motion_field:MF_final}).
\end{enumerate}

\input{figure_masks}

\paragraph{Motion-adapted Volume Generation}
The breathing state at each X-ray projection is modeled as a sinusoidal function with random perturbations to simulate breathing irregularities:

\begin{equation}
    s_i = \sin\left(\pi \frac{i \cdot T_p + \epsilon_i}{T_{hc}}\right)
    \label{eq:motion}
\end{equation}

where $i$ is the projection index, $T_p$ is the acquisition time per projection, $T_{hc}$ is the half-cycle breathing period, and $\epsilon_i \sim \mathcal{N}(0, (20 \text{ ms})^2)$ introduces temporal variability in breathing rhythm.

The displacement field at projection $i$ is computed by scaling the motion field with the breathing state and maximum amplitude:

\begin{equation}
    \mathbf{D}(\mathbf{x}, i) = A_{max} \cdot \mathbf{M}(\mathbf{x}, s_i)
    \label{eq:deformation_field}
\end{equation}

The breathing amplitude $A_{max}$ corresponds to ±$A_{max}$ peak displacement, with +$A_{max}$ representing maximum inhalation and -$A_{max}$ representing maximum exhalation. Motion-adapted volumes are generated by resampling the contrast-free CT at positions $\mathbf{x} + \mathbf{D}(\mathbf{x}, i)$ using trilinear interpolation.

\subsection{Dynamic X-ray Projection}
\label{sec:drr_recon}
Our X-ray imaging simulation framework was implemented in CUDA (v12.8) and CuPy, inspired by the DeepDRR package \cite{DeepDRR2018}, to enable high-performance computation of line integrals through ray casting. The framework incorporated motion-adapted projection capabilities through dedicated CUDA kernels that dynamically resampled the CT volume according to time-varying deformation fields during each projection acquisition.

\paragraph{Contrast Media Removal}
Patients often received contrast agent prior to fan-beam \gls{CT} acquisition for radiotherapy to visualize organs for improved delineation. To simulate \gls{CBCT} images without contrast enhancement, visible contrast media in the surrogate structure was suppressed. First, a binary mask $M_{binary}(\mathbf{x})$ identified voxels within a predefined segmentation exceeding 50 HU. This mask was convolved with a Gaussian kernel ($\sigma$ = 1) to create a fuzzy mask $M(\mathbf{x})$. For each voxel, a noise factor $n(\mathbf{x})$ was sampled from a Gaussian distribution to create smooth intensity transitions:

\begin{equation}
    n(\mathbf{x}) \sim \mathcal{N}(1, 0.02^2)
\end{equation}

The contrast-free CT intensity was computed as:

\begin{equation}
    I_{cm-free}(\mathbf{x}) = I_{CT}(\mathbf{x}) - 0.92 \cdot I_{CT}(\mathbf{x}) \cdot M(\mathbf{x}) \cdot n(\mathbf{x})
\end{equation}

where the factor 0.92 was empirically determined to best replicate the appearance of non-contrast \gls{CT} acquisitions. The contrast-free \gls{CT} was used for all subsequent processing steps.

\paragraph{Scanner Geometry Configuration} The simulation framework is able to replicate cone-beam scanner geometry through configurable acquisition parameters. These included detector dimensions and pixel spacing, source-to-axis and source-to-detector distances, angular sampling density and total rotation arc, as well as detector offset for extended \gls{FOV} configurations. The rotation center can be positioned arbitrarily within the imaging volume, enabling simulation of both standard and offset detector geometries used across different clinical systems. To best represent clinical setup, the isocenter used for the treatment plan can be used as the rotation center. All scanner specific settings can be found in \autoref{tab:parameter}\\

\paragraph{Motion-Adapted Projection Generation}
For each projection angle $\theta_i$ (where $i = 1, \ldots, n$), the \gls{CT} volume was dynamically resampled according to the time-varying deformation field. The breathing state $s_i$ was computed using \autoref{eq:motion}, and the displacement field (\autoref{eq:deformation_field}) was applied through trilinear interpolation to generate a motion-adapted volume. Ray casting was then performed by defining rays from the X-ray source to individual detector pixels and accumulating attenuation coefficients along each ray path. This dynamic resampling approach ensured that respiratory motion artifacts were properly incorporated into the simulation, providing realistic representation of clinical imaging scenarios where patient motion occurs during the acquisition period.

\paragraph{Physics-Based Imaging Model}
Prior to projection, CT \gls{HU} values were converted to linear attenuation coefficients using \\ $\mu =  {\mu_{water}\cdot (\text{HU}+1000)}/{1000}$ with $\mu_{water} =$ 0.018 mm$^{-1}$.

The incident photon distribution at the detector was modeled as:
\begin{equation}
    I_0(i,j) = BP(i,j) \cdot mAs \cdot \Phi \cdot p_{size}^2
    \label{eq:incident_photons}
\end{equation}
where $BP(i,j)$ represents a 2D Gaussian beam profile fitted to measured bowtie filter characteristics from the clinical scanner. The beam profile was constructed by computing the mean profile along the longitudinal detector axis and fitting a 1D Gaussian function using scipy's curve fitting. This profile was replicated across the orthogonal axis to create the 2D distribution. The tube current-time product mAs was determined from real measurements (\autoref{sec:patient_cohort}), $\Phi$ is the photon fluence per unit dose, and $p_{size}$ is the detector pixel pitch (see \autoref{tab:parameter}).

Following the Beer-Lambert law, the attenuation projection $P(i,j)$ was computed through ray casting, and primary transmission was calculated as:
\begin{equation}
    T_{primary}(i,j) = \exp(-P(i,j))
\end{equation}
yielding primary photon counts:
\begin{equation}
    N_{primary}(i,j) = T_{primary}(i,j) \cdot I_0(i,j)
\end{equation}

\paragraph{Scatter Radiation Model}
We implemented a heuristic scatter model that accounts for the spatial distribution of attenuating tissue. First, an attenuation mask was defined to identify regions of significant X-ray absorption:
\begin{equation}
    \Omega(i,j) = \begin{cases}
        1 & \text{if } T_{primary}(i,j) < 0.5 \\
        0 & \text{otherwise}
    \end{cases}
\end{equation}

For each projection, scatter photon counts were estimated using the fifth percentile of primary counts across the detector as a reference level:
\begin{equation}
    N_{scatter}(i,j) = SPR \cdot P_5(N_{primary}) \cdot \Omega(i,j)
\end{equation}
where $P_5(N_{primary})$ denotes the fifth percentile computed over all detector pixels in the current projection, and the scatter-to-primary ratio $SPR$ was empirically set to match clinical \gls{CBCT} appearance (\autoref{tab:parameter}). 

The multiplication by primary counts ensures that scatter intensity scales appropriately with tissue density.

\paragraph{Noise Model and Projection Finalization}
Photon detection noise was modeled using Poisson statistics:
\begin{equation}
    N_{detected}(i,j) \sim \text{Poisson}(N_{primary}(i,j) + N_{scatter}(i,j))
\end{equation}

The detected photon counts were then normalized by the incident photon distribution, corrected for the detector saturation and converted to logarithmic attenuation space to generate the final noisy projection:

\begin{equation}
P_{noisy}(i,j) = -\ln\left(c_{sat} \cdot \frac{N_{detected}(i,j)}{I_0(i,j)}\right)
\end{equation}

The saturation constant $c_{sat}$ models the detector's limited dynamic range (\autoref{tab:parameter}), with the transmission ratio $ c_{sat}\cdot N_{detected}/I_0$ clamped to [10\textsuperscript{-6}, 1.0]
 before applying the logarithm to ensure numerical stability. $c_{sat}$ was selected experimentally according to represent best clinical settings as well as verified via an ablation study (see \ref{app1}).

\subsection{CBCT Reconstruction}
\label{sec:recon}
Simulated \gls{CBCT} volumes were reconstructed from the simulated projections using the Feldkamp-Davis-Kress (FDK) algorithm implemented in the Reconstruction Toolkit (RTK) package version 2.7.0 \cite{Rit2014}. The reconstruction used Hann window with cutoff frequency of 0.9 and Truncation correction set to 0.05. The reconstructed volume parameters (dimensions and resolution) can be found in \autoref{tab:parameter}. The rotation axis was positioned at the treatment planning isocenter to ensure geometric correspondence with the clinical acquisition geometry.

Following reconstruction, gray values were converted to \gls{HU} using the conversion formula $\mu\times65536-1024$. Regions outside the field of view were identified using a cone-shaped mask determined by the scanner geometry and set to -1024 HU.

\paragraph{Reference CT Preparation}
For each \gls{sCBCT}, the corresponding fan-beam \gls{CT} was resampled to the same reference space ($410 \times 410 \times 66$ voxels, 1 mm $\times$ 1 mm $\times$ 4 mm) and truncated using the identical \gls{FOV} mask. This created perfectly geometrically aligned \gls{sCBCT}/\gls{CT} pairs.

%% file: figure_masks.tex
\begin{figure}[!ht]
\centering
\begin{subfigure}[b]{0.45\textwidth}
    \includegraphics[width=\textwidth]{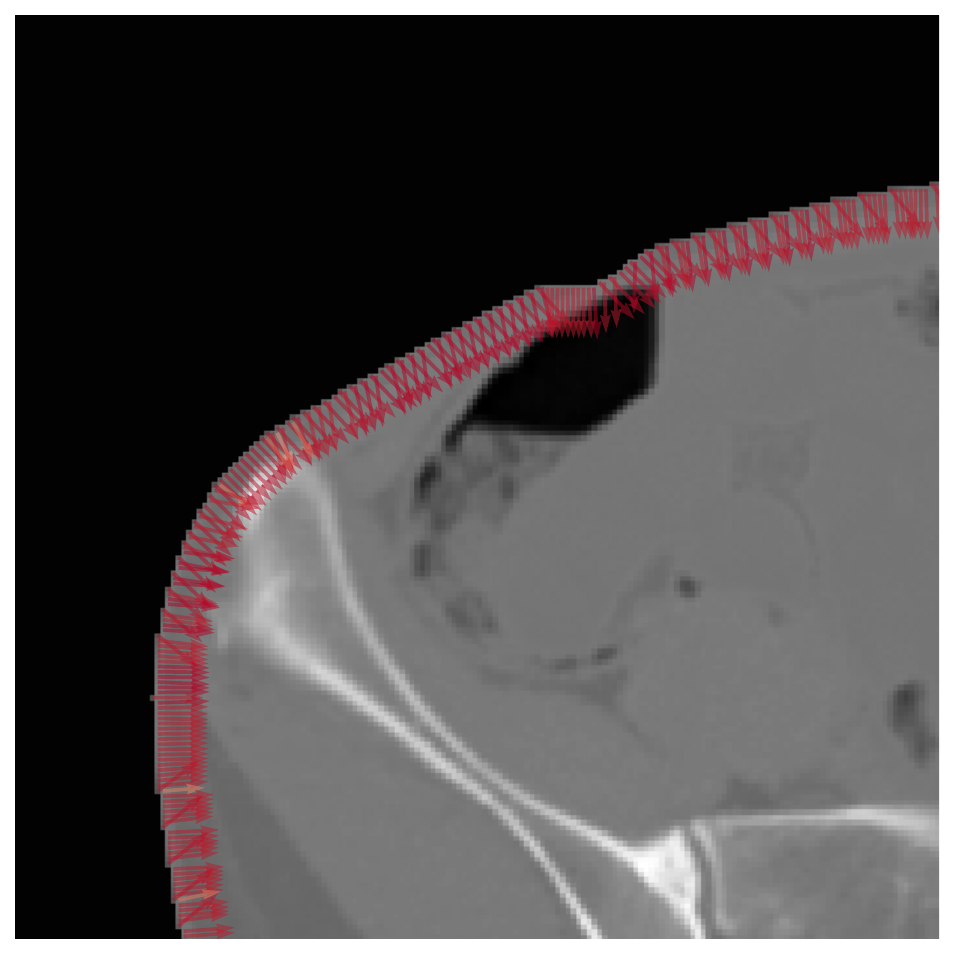}
    \caption{$V_{s}$}
    \label{fig:motion_field:MF_strong_border_gradients}
\end{subfigure}
\begin{subfigure}[b]{0.45\textwidth}
    \includegraphics[width=\textwidth]{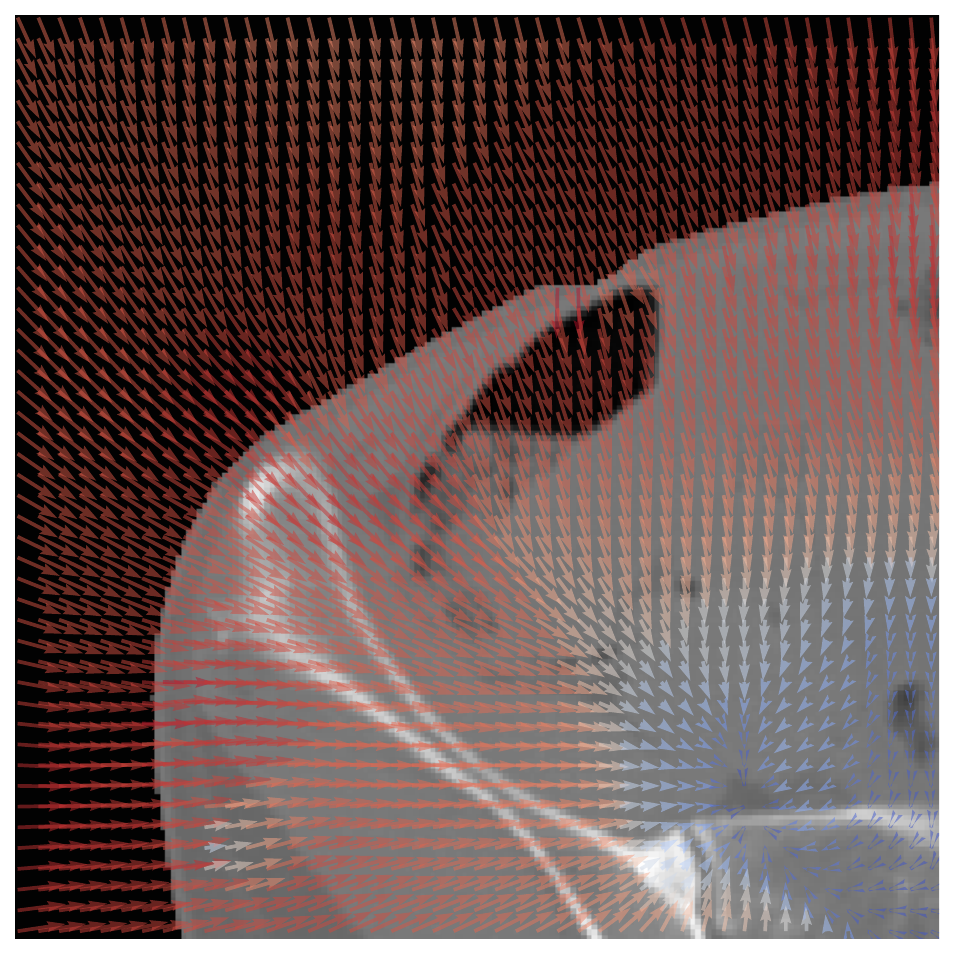}
    \caption{$V_{d}$}
    \label{fig:motion_field:MF_propagated_gradients1}
\end{subfigure}
\vspace{0.4cm}

\begin{subfigure}[b]{0.45\textwidth}
    \includegraphics[width=\textwidth]{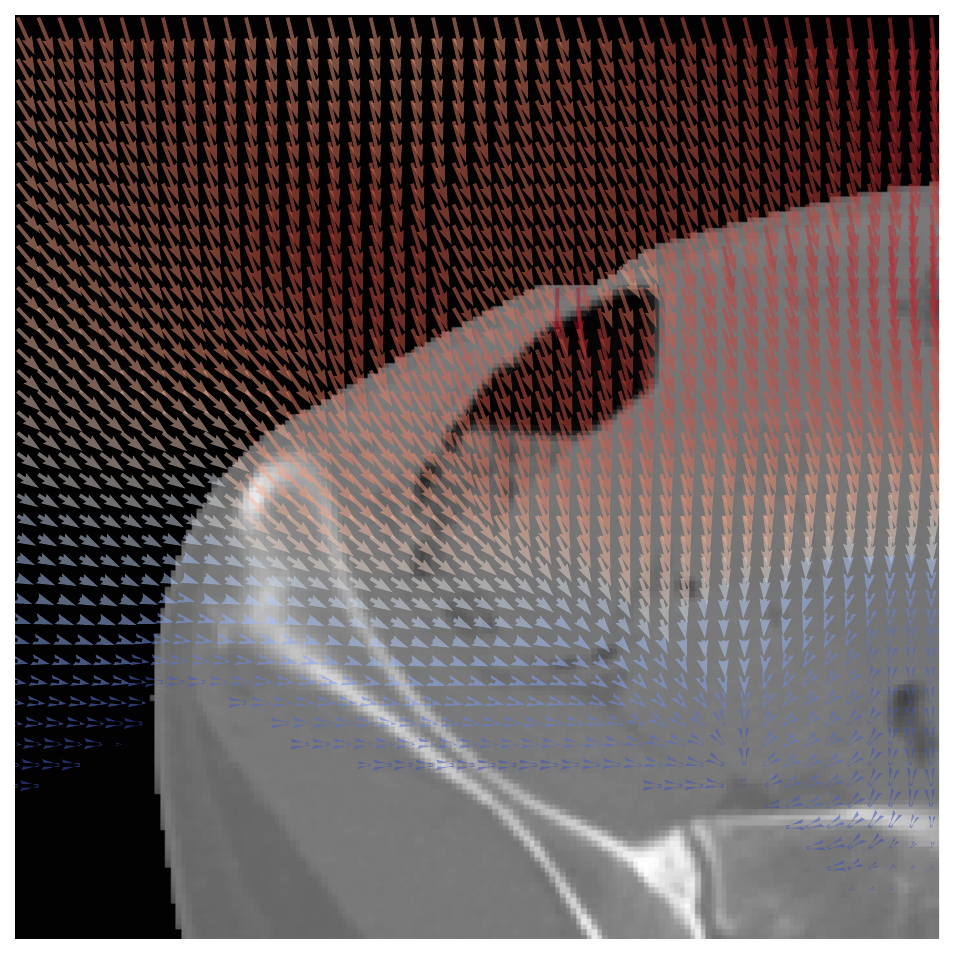}
    \caption{$V_{u}$}
    \label{fig:motion_field:MF_propagated_gradients2}
\end{subfigure}
\begin{subfigure}[b]{0.45\textwidth}
    \includegraphics[width=\textwidth]{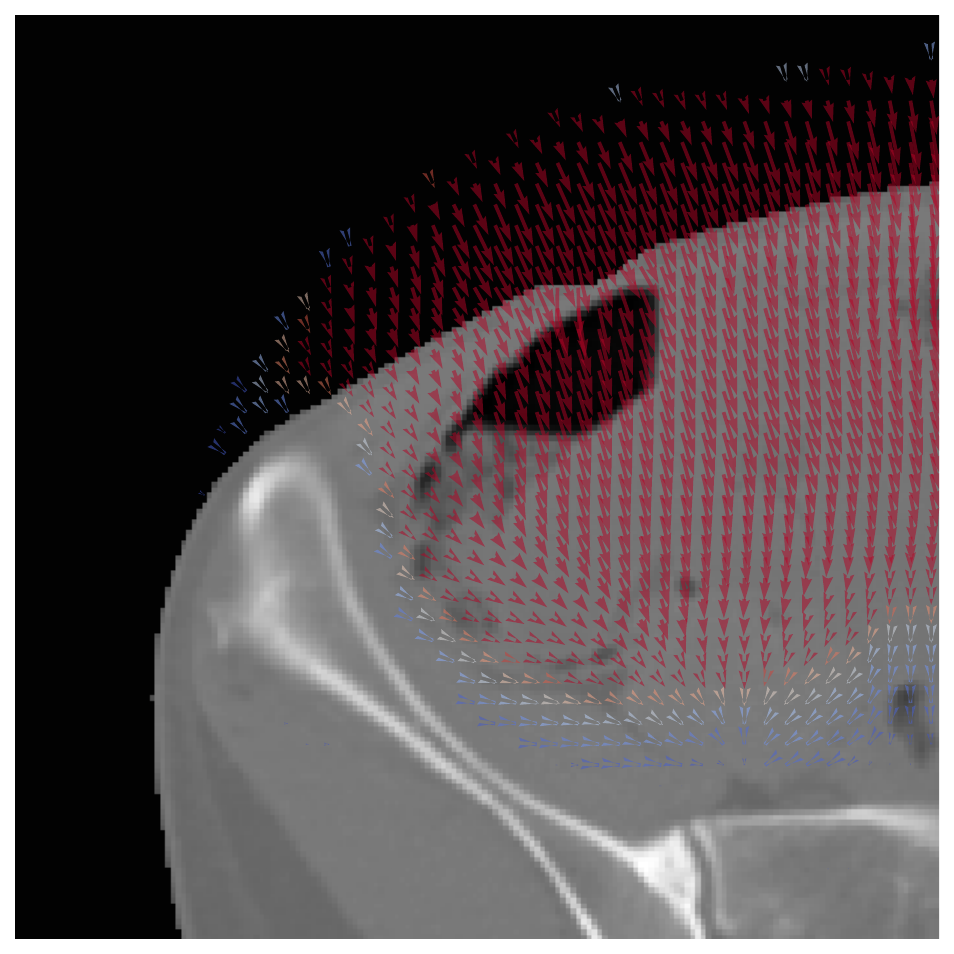}
    \caption{$V_{f}$}
    \label{fig:motion_field:MF_final}
\end{subfigure}

\caption{Motion vector field generation (x and y component of 3D motion vector visualized as 2D vector overlay on a central axial slice of test patient volume):
the sparse motion vector field (\autoref{fig:motion_field:MF_strong_border_gradients}) encodes strong gradient vectors in border voxels, the dense motion vector field (\autoref{fig:motion_field:MF_propagated_gradients1}) has a motion vector estimate for every voxel position, the updated motion vector field (\autoref{fig:motion_field:MF_propagated_gradients2}) attenuates motion vectors that deviate from the desired main motion direction, and the final motion vector field (\autoref{fig:motion_field:MF_final}) only keeps motion vectors in the extended proximity of the reference structure and eliminating motion vectors at voxels representing bone or voxels behind bones.}

\label{fig:motion_field:vectorfields}
\end{figure}

%% file: experiments.tex
\subsection{Patient cohort}
\label{sec:patient_cohort}
\textit{Clinical dataset.} Imaging data from 395 gynecological patients treated at the Department of Radiation Oncology at the Medical University of Vienna between 2022 and 2024 were included. Each patient dataset encompassed \gls{CT} and \gls{CBCT} images as well as DICOM structure files including organs and target volume segmentations, and the treatment plan. For patient inclusion, bowel bag delineation was required as surrogate structure elaborated in more detail in \autoref{ch:respiratory_motion}. All \gls{CBCT} scans were acquired using an X-ray volumetric imaging (XVI) system (Elekta AB, Stockholm, Sweden). The \gls{CT} scans were acquired using a Siemens Somatom Definition AS scanner (Siemens Healthineers, Germany). This study was approved by the institutional ethics committee (approval number: 1359/2025).\\
All 395 \gls{CT} images were used to generate \gls{sCBCT}/\gls{CT} pairs using the isocenter extracted from the radiotherapy treatment plan to position the patient according to clinical treatment geometry. For our experiments, we replicated the Elekta XVI system in medium \gls{FOV} configuration using the settings defined in \autoref{tab:parameter}. For evaluating model performance on \gls{CBCT} data, a subset of 180 patients with \gls{CBCT} scans was selected as training data. These \gls{CBCT}/\gls{CT} pairs required spatial alignment. The clinically approved \gls{CBCT} positioning was used as the initial alignment, followed by deformable registration performed with the Elastix registration tool \cite{Klein2010}. For deformable registration, we employed the IMPACT loss metric \cite{Boussot2025a}, which uses a pretrained deep learning model (Total Segmentator, M730) as a feature extractor. The metric compares feature maps from two network layers using L2 distance. Registration parameters included a grid spacing of 16 mm and two resolution layers. An additional 30 patients with \gls{CBCT} scans were designated as a held-out test set and underwent the same registration procedure.\\
\textit{SynthRAD dataset.} Additionally, the SynthRAD Challenge 2023 dataset was included \cite{SynthRAD2023}. All available \gls{CT} scans in the pelvic region were reviewed, and unsuitable images (\gls{CT} scans with insufficient \gls{FOV} for motion simulation) were excluded, resulting in 349 patients from which \gls{sCBCT}/\gls{CT} pairs were generated. Since the SynthRAD dataset does not include bowel bag delineations required for motion simulation (\autoref{ch:respiratory_motion}), we trained a segmentation model to generate these labels. An nnUNet model with medium sized ResNet encoder and default configuration was trained on the clinical dataset, where bowel bag contours were available from clinical practice. This model was then applied to segment bowel bag structures in the SynthRAD \gls{CT} images. All predicted segmentations were visually reviewed and manually corrected by a data scientist to ensure anatomical accuracy before use in the simulation pipeline. The dataset includes a variety of gender and tumor types. As the isocenter was not available for these patients, the image center was used as the rotation center for physics simulation. 180 rigidly registered \gls{CBCT}/\gls{CT} pairs were used as provided by the challenge organizers. The challenge validation set of 30 patients in the pelvic region was used as an additional held-out test set for our study and is referred to as such throughout this manuscript.\\
Table \ref{tab:datasets} summarizes the dataset composition for both simulated and real data configurations.\\
\begin{table}[!htbp]
\centering
\begin{tabular}{lccc}
\toprule
&simulated \gls{CBCT} &  \multicolumn{2}{c}{real \gls{CBCT}}\\
\midrule
&training/validation & training/validation& test\\
\midrule
SynthRAD    & 349 & 180 & 30\\
Clinical  & 395  & 180 & 30 \\
\bottomrule
\end{tabular}
\caption{Dataset composition showing sample sizes for the SynthRAD Challenge dataset and clinical data, all in the pelvic region. Training data was split into five folds for cross-validation following nnUNet conventions. All planning \gls{CT} images were used to generate \gls{sCBCT}/\gls{CT} pairs, while real \gls{CBCT}/\gls{CT} pairs represent subsets with acquired \gls{CBCT} scans.}
\label{tab:datasets}
\end{table}

\subsection{Model training}
For all experiments, Nvidia A100 GPUs of the high performance cluster at the Medical University of Vienna were utilized. The regression performance was analyzed in the context of \gls{CBCT}-to-\gls{CT} translation. For all runs, a modified nnUNet regression trainer was implemented and used with ResNet encoder configuration medium size, deep supervision loss enabled and L1 loss \cite{Isensee2021}. The input \gls{CBCT} images were normalized using Z-scoring and the \gls{CT} labels were normalized using a global normalization where the mean and standard deviation over the full training dataset were recorded and used during training and inference for normalization and renormalization.\\
Five-fold cross-validation was performed for both the SynthRAD and clinical datasets. The models were trained for 1000 epochs with a learning rate of 0.01 and the SGD optimizer including Nesterov momentum set to 0.99. Three training configurations were evaluated: (1) \Ms: trained from scratch on synthetic \gls{sCBCT}/\gls{CT} pairs generated from real \gls{CT} images using physics simulation (\autoref{sec:framework}), (2) \Mr: trained from scratch on aligned \gls{CBCT}/\gls{CT} pairs, and (3) \Mf: initialized with weights pretrained on synthetic data, then finetuned on real \gls{CBCT}/\gls{CT} pairs.\\
For \Mf, a model was first pretrained on all available simulated data to create an initial checkpoint. Subsequently, the encoder portion of this pretrained model was frozen during finetuning, while the decoder was trained for 50 epochs with a reduced learning rate of 0.001 on the real data from each fold. All other hyperparameters remained identical to the initial training.\\
For evaluation, predictions from all five folds were ensembled by averaging. Performance metrics were reported on the separate held-out test sets consisting of 30 cases each from the SynthRAD and clinical datasets.\\

\subsection{Evaluation}

The realism of \gls{sCBCT} images was evaluated using \gls{FID} and \gls{MMD} metrics. \gls{FID} was computed using an Inception v3 model as the feature extractor, while \gls{MMD} utilized ViT-L/16 (both accessed from torchvision v0.25.0). 3D images were converted to 2D transversal slices for metric computation. Metrics were calculated separately for simulated data generated from the SynthRAD and clinical datasets. As reference baselines, we also report \gls{FID} and \gls{MMD} metrics (1) between SynthRAD and clinical data, and (2)  between two equal splittings of the same dataset (called from now on internal \gls{FID} and \gls{MMD} metrics), to provide expected value ranges for interpretation.\\
For the \gls{CBCT}-to-\gls{CT} translation task, we evaluated \gls{MAE}, \gls{SSIM}, and \gls{PSNR}. To test if the output preserves geometric correspondence with the actual geometric anatomy, additionally \gls{NMI} and \gls{CC} between predicted \gls{sCT} and the input \gls{CBCT} were computed. These metrics measure whether the output preserves geometric correspondence with the actual imaged anatomy, independent of ground truth quality. All metrics were computed within an outline mask as defined for the SynthRAD Challenge 2023 to ensure comparability \cite{SynthRAD2023}. The mask was defined by an Otsu threshold with different morphological operations. Statistical comparisons between model configurations were performed using the Wilcoxon signed-rank test. To account for multiple comparisons across the five metrics, Bonferroni correction was applied. Effect sizes were quantified using Cohen's d with interpretation thresholds of $|d| < 0.2$ (negligible), $0.2 \leq |d| < 0.5$ (small), $0.5 \leq |d| < 0.8$ (medium), and $|d| \geq 0.8$ (large).\\
To complement quantitative metrics and mitigate potential selection bias, a \gls{IQS} analysis was conducted by five independent observers (Data scientists, Medical physicists, Radiation technologists). Observers evaluated clinical (n=30) and SynthRAD (n=30) testset predictions from \Ms and \Mr using a custom visualization tool that displayed the original \gls{CBCT} alongside both reconstructed \gls{CT} volumes with synchronized crosshair navigation. Each observer scored both reconstructions on a scale from 0 (very poor) to 5 (excellent) based on anatomical accuracy, image quality, artifact severity, and overall reconstruction fidelity. Additionally, observers indicated their preference: (1) prefer \Ms, (2) prefer \Mr, or (3) no preference. The inter-rater consistency was evaluated using Randolph's Kappa test as well as \gls{ICC} for analyzing consistency (two-way mixed effects, average rater). The \gls{IQS} was designed to validate whether geometric alignment corresponds to clinical perception of quality. Therefore, we performed a correlation analysis between all metrics and the mean \gls{IQS} over all observers using Spearman's Rank correlation analysis reporting p-value and the \gls{CI95}. Additionally, a Steiger's Z test was performed comparing the correlation dependency (\eg $\rho$(\gls{NMI}, \gls{IQS})$>$$\rho$(\gls{MAE}, \gls{IQS})).\\
All statistical analyses were performed using Python 3.12.3 (pingouin v0.5.5).

%% file: results.tex
\section{Results}
\label{res}
\subsection{Simulation Realism}
\autoref{tab:FID} presents the \gls{FID} and \gls{MMD} scores evaluating simulated \gls{CBCT} quality. For the SynthRAD dataset, simulated data achieved comparable distribution metrics to the data split comparison of the same dataset (\gls{FID}: 63.4 vs 52.73; \gls{MMD}: 0.69 vs 1.07), indicating that simulated images captured the statistical properties of real \gls{CBCT} data. The clinical dataset demonstrated lower \gls{FID} scores overall (synthetic: 22.46; internal: 7.73), suggesting more homogeneous imaging characteristics. While the simulated-to-real gap appeared larger for clinical data, the cross-dataset comparison between real SynthRAD and real clinical data (\gls{FID}: 42.78, \gls{MMD}: 0.85) demonstrated that distribution differences across datasets can be substantial even among real acquisitions. Further, computing the intensity metrics on the \gls{sCBCT}/\gls{CT} pair of the validation sets resulted in 6.7 HU (\gls{MAE}), 47.1 dB (\gls{PSNR}) and 0.99 (\gls{SSIM}) for all 395 Patients of the clinical dataset. \gls{sCBCT} image generation took approximately \SI{10}{\second} per 10M voxel dataset.\\

\begin{table}[htbp!]
    \centering
    \begin{tabular}{cccc}
    \toprule
    Dataset & Comparison & FID $\downarrow$ & MMD $\downarrow$ \\
    \midrule
     \multirow{2}{*}{SynthRAD} &  simulated vs. real & 63.43 & 0.69\\
       & internal (real vs. real) & 52.73 & 1.07\\
    \midrule
         \multirow{2}{*}{Clinical} &  simulated vs. real & 22.46 & 0.68\\
       & internal (real vs. real) & 7.73 & 0.06\\
    \midrule
    \multicolumn{2}{c}{SynthRAD vs. Clinical (both real)} & 42.78 & 0.84\\
    \bottomrule
    \end{tabular}
    \caption{\Gls{FID} and \gls{MMD} scores evaluating simulated \gls{CBCT} quality. Lower values indicate more similar distributions. "Simulated vs. real" compares generated images to real acquisitions within each dataset. "Internal" compares two subsets of real data within each dataset. The cross-dataset comparison shows distribution differences between real SynthRAD and clinical data.}
    \label{tab:FID}
\end{table}

\subsection{Model Performance}
\begin{figure}
    \centering
    \includegraphics[width=\linewidth]{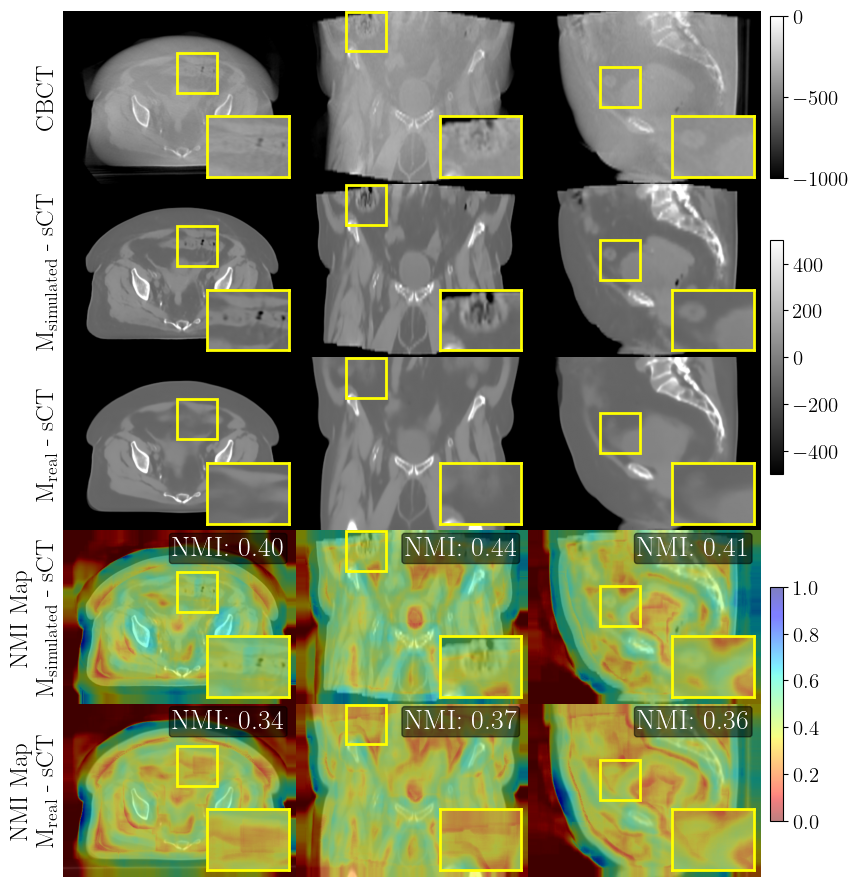}
    \caption{Qualitative comparison of geometric alignment between \Ms (second row) and \Mr (third row) visualized using local \gls{NMI} heatmaps for SynthRAD test case 2PA020. Higher values (blue regions) indicate stronger local alignment with the source \gls{CBCT}. \Ms demonstrates superior alignment throughout the volume, particularly in anatomically complex regions.}
    \label{fig:nmi_comparison}
\end{figure}

\autoref{tab:regression_multiset_comparison} presents comprehensive evaluation results across all training and testing configurations. The results demonstrated that models trained on synthetic data (\Ms) consistently achieved superior or equivalent geometric alignment compared to models trained on real data (\Mr), as measured by \gls{NMI} and \gls{CC}.\\
A systematic discrepancy emerged between intensity-based and geometric metrics. \Mr achieved lower \gls{MAE} (34.76 vs 48.48 HU) when evaluated against ground truth, yet \Ms demonstrated better geometric alignment with the input \gls{CBCT} (\gls{NMI}: 0.30 vs 0.28, \gls{CC}: 0.98 vs 0.97). The effect was most pronounced in cross-dataset evaluation (models were trained on rigidly aligned data but tested on deformable registered data), where \Ms improved \gls{NMI} by 41\% relative to \Mr (0.31 vs 0.22), indicating that registration methodology differences between training and test amplify bias effects.\\
Performance patterns varied by train-test dataset configuration (\autoref{tab:regression_multiset_comparison}). When both used the same registration methodology, \Mr achieved lower \gls{MAE}; when registration methodologies differed, \Ms achieved lower \gls{MAE}. For models trained on rigidly registered SynthRAD data and tested on deformably registered clinical data, \Ms improved \gls{MAE} by 32.68 HU (39\% reduction), \gls{PSNR} by 4.60 dB, and \gls{SSIM} by 0.06 compared to \Mr.
Finetuning synthetic-pretrained models on real data (\Mf) yielded performance approaching that of \Mr, suggesting that synthetic pretraining provides useful initialization that can be refined with limited real data.\\
\autoref{fig:nmi_comparison} provides qualitative visualization of the geometric alignment advantage, showing local \gls{NMI} heatmaps for \Ms and \Mr predictions on SynthRAD test case 2PA020. \Ms demonstrated consistently higher alignment throughout the anatomy, particularly in anatomically complex regions.\\

\input{regression_results}

\subsection{Observer assessment}

The \gls{IQS} conducted with five independent observers (total evaluations across all test cases) strongly favored outputs from \Ms. Observers preferred \Ms in 87\% of cases, with an average quality rating of 3.94 compared to 2.04 for \Mr. No clear preference was expressed in 8.8\% of cases, while \Mr was preferred in only 4.2\% of cases. Preferences had substantial agreement (Clinical and SynthRad: $\kappa$=0.62). Inter-rater reliability metrics (\autoref{tab:rater_stats}) show poor values for both datasets and models comparing the observer scores. For both datasets, \Ms was more consistent in the rating than \Mr.\\
\noindent\begin{table}[!h]
    \centering
    \begin{tabular}{cccc}
    \toprule
    \textbf{Dataset} & \textbf{Model type} & \textbf{$\kappa$} &  \textbf{ICC}\\
    \midrule
   \multirow{2}{*}{Clinical}   &\Ms &   0.18 &0.30\\
      &\Mr &  0.12 &0.19\\
   \multirow{2}{*}{SynthRAD}   &\Ms &   0.19 &0.45\\
      &\Mr &  0.05 &0.24\\
         \bottomrule
    \end{tabular}
    \captionof{table}{Comparison of inter-rater reliability metrics, Randolph's kappa ($\kappa$) and the intraclass correlation coefficient model ICC, for ratings derived from \Ms and \Mr on SynthRAD and Clinical datasets.}
    \label{tab:rater_stats}
\end{table}
\hfill
\begin{table}[!h]
    \centering

    \begin{tabular}{ccccc}
    \toprule
    \textbf{Metric} & \textbf{Dataset} & $\mathbf{\rho}$ & \textbf{CI95\%} &  \textbf{p-value}\\
    \midrule
   \multirow{3}{*}{MAE} & Clinical & 0.57 & [0.37, 0.71] & $<$0.001\\
      & SynthRAD & -0.10 & [-0.36, 0.19] & 0.45\\
      & Combined & 0.02 & [-0.16, 0.22] & 0.80\\
   \midrule
   \multirow{3}{*}{PSNR} & Clinical & -0.46 & [-0.63, -0.25] & $<$0.001\\
      & SynthRAD & 0.16 & [-0.15, 0.41] & 0.23\\
      & Combined & 0.03 & [-0.17, 0.22] & 0.75\\
   \midrule
   \multirow{3}{*}{SSIM} & Clinical & -0.59 & [-0.72, -0.40] & $<$0.001\\
      & SynthRAD & 0.10 & [-0.16, 0.36] & 0.43\\
      & Combined & -0.05 & [-0.24, 0.14] & 0.58\\
   \midrule
   \multirow{3}{*}{NMI} & Clinical & 0.29 & [0.04, 0.52] & 0.02\\
      & SynthRAD & 0.31 & [0.06, 0.54] & 0.02\\
      & Combined & 0.31 & [0.14, 0.45] & $<$0.001\\
   \midrule
   \multirow{3}{*}{CC} & Clinical & 0.06 & [-0.20, 0.31] & 0.65\\
      & SynthRAD & -0.06 & [-0.28, 0.18] & 0.67\\
      & Combined & 0.05 & [-0.14, 0.22] & 0.62\\
         \bottomrule
    \end{tabular}
    
    \caption{Spearman correlation between image quality metrics and clinical assessment (\gls{IQS}), stratified by dataset and combined. n=60 per dataset, n=120 combined.}
    \label{tab:correlation}
\end{table}

\begin{figure}[]
    \centering
    \includegraphics[width=0.85\linewidth]{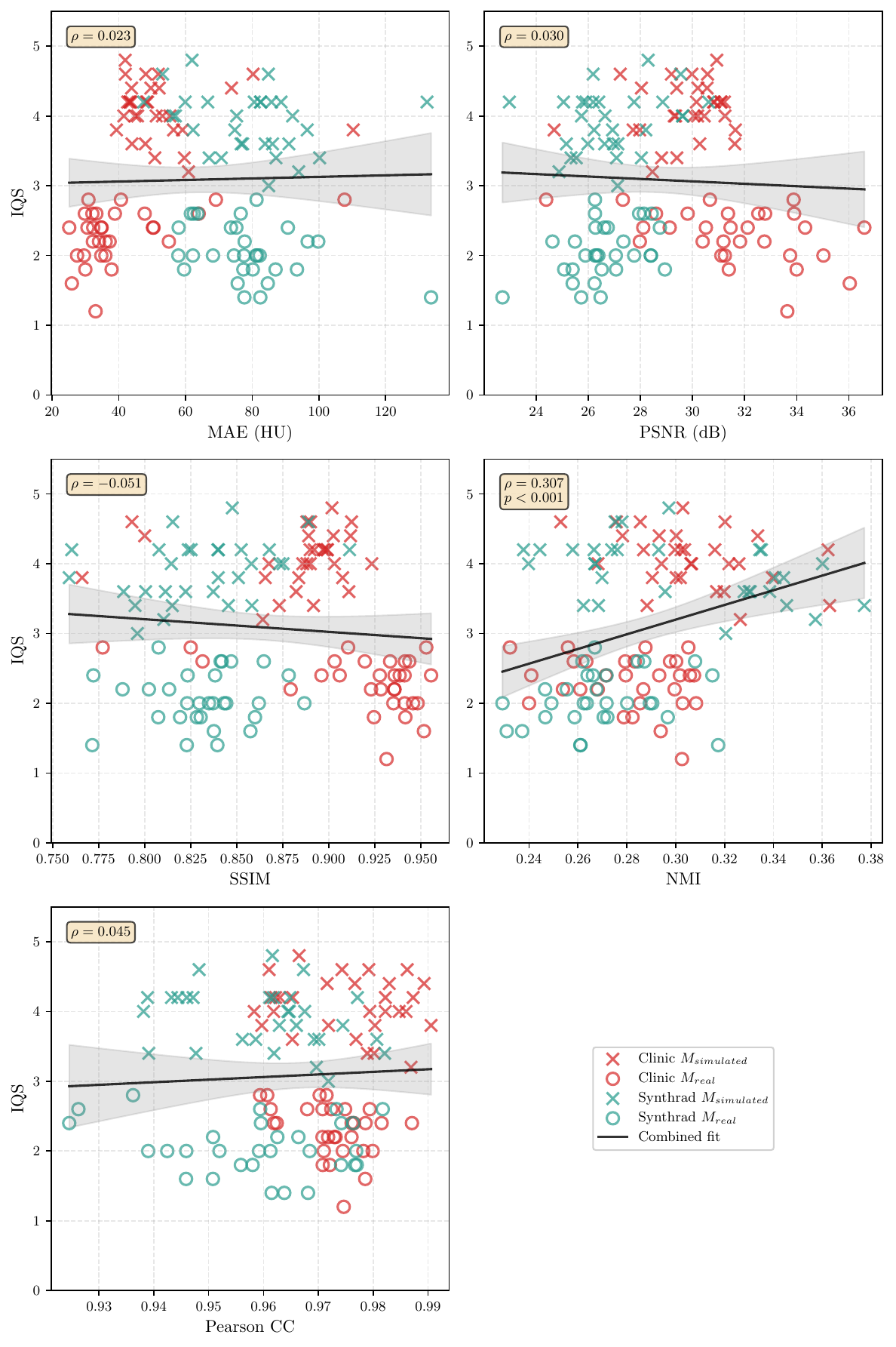}
    \caption{Relationship between quantitative metrics and clinical visual assessment scores. Green and red markers show samples from SynthRAD and Clinical datasets, respectively. Circles indicate \Mr and crosses show \Ms samples. Note the inverted \gls{MAE}-\gls{IQS} relationship for Clinical data (red markers) compared to SynthRAD (green markers).}
    \label{fig:iqs_correlation}
\end{figure}

The computed Spearman correlations between per-case metric values and mean \gls{IQS} across observers  can be found in \autoref{tab:correlation}. A divergence emerged between datasets with different registration methodologies which can be seen in \autoref{fig:iqs_correlation}.\\
For the clinical dataset (deformable registration), intensity metrics exhibited strong but inverted correlations with clinical assessment. Higher \gls{MAE} correlated with higher \gls{IQS} ($\rho$ = 0.57, p $<$ 0.001), while higher \gls{PSNR} and \gls{SSIM} correlated with lower \gls{IQS} ($\rho$ = -0.46 and $\rho$ = -0.59, respectively, both p $<$ 0.001). This inversion indicated that outputs achieving better intensity metrics were systematically rated as clinically inferior by observers.\\
For the SynthRAD dataset (rigid registration), intensity metrics showed no significant correlation with \gls{IQS} (all p $>$ 0.2).\\
In contrast, \gls{NMI} demonstrated consistent positive correlation with \gls{IQS} across both datasets (clinical: $\rho$ = 0.29, p = 0.02; SynthRAD: $\rho$ = 0.31, p = 0.02), confirming its validity as a predictor of clinical preference independent of ground truth quality. \gls{CC} showed no significant correlation in either dataset.\\
When datasets were combined, intensity metric correlations were small and not significant and only \gls{NMI} showed moderate correlation (p$<$0.001). Steiger's Z tests confirmed that \gls{NMI} predicted \gls{IQS} significantly better than \gls{MAE} (Z = 2.27, p = 0.02) and \gls{SSIM} (Z = 2.53, p = 0.01) in the combined analysis.

%% file: regression_results.tex
\begin{table}[htbp]
\centering
\caption{Regression Performance: Clinical vs SynthRAD for all Training and Testing dataset combinations. Values depicted are the median with the 25\textsuperscript{th} and the 75\textsuperscript{th} percentiles in brackets. Best performing configuration is marked in bold per metric and combination (training and test dataset). Asterisk mark significant results and ES gives Cohen's D effect size to \Ms. As reference, SynthRAD challenge winner metrics for the pelvic dataset is also shown.}
\label{tab:regression_multiset_comparison}
\begin{adjustbox}{max width=\textwidth}
\begin{tabular}{cclcccccc}
\toprule
\thead{\textbf{Test}\\ \textbf{Dataset}} & \thead{\textbf{Training}\\ \textbf{Data}} & \thead{\textbf{Model}\\ \textbf{Type}} & &\thead{\textbf{MAE (HU) $\downarrow$}} & \thead{\textbf{PSNR (dB) $\uparrow$}} & \thead{\textbf{SSIM $\uparrow$}} & \thead{\textbf{NMI $\uparrow$}} & \thead{\textbf{CC $\uparrow$}} \\
\midrule
\multirow{12}{*}{\rotatebox{90}{Clinical}} & \multirow{6}{*}{\rotatebox{90}{Clinical}} & \multirow{2}{*}{\Ms} & median & 48.48 & 30.17 & 0.89 & \textbf{0.30} & \textbf{0.98} \\
 &  &  & IQR & (43.79, 54.81) & (29.22, 30.83) & (0.88, 0.90) & (0.29, 0.32) & (0.97, 0.98) \\
 &  & \multirow{2}{*}{\Mr} & median / ES & \textbf{34.76$^{*}$} / 0.80 & \textbf{31.43$^{*}$} / -0.84 & \textbf{0.93$^{*}$} / -0.95 & 0.28$^{*}$ / 1.08 & 0.97$^{*}$ / 0.29 \\
 &  &  & IQR & (31.23, 40.22) & (30.42, 33.72) & (0.91, 0.94) & (0.26, 0.30) & (0.97, 0.98) \\
 &  & \multirow{2}{*}{\Mf} & median / ES & 37.70$^{*}$ / 0.70 & 31.40$^{*}$ / -0.81 & 0.93$^{*}$ / -0.86 & 0.29$^{*}$ / 0.90 & 0.98$^{*}$ / -0.06 \\
 &  &  & IQR & (33.39, 42.04) & (30.34, 33.43) & (0.91, 0.94) & (0.27, 0.30) & (0.97, 0.98) \\
\cmidrule{2-9}
 & \multirow{6}{*}{\rotatebox{90}{SynthRAD}} & \multirow{2}{*}{\Ms} & median & \textbf{50.56} & \textbf{29.96} & \textbf{0.89} & \textbf{0.31} & \textbf{0.98} \\
 &  &  & IQR & (44.69, 57.11) & (29.15, 30.66) & (0.88, 0.90) & (0.29, 0.32) & (0.96, 0.98) \\
 &  & \multirow{2}{*}{\Mr} & median / ES & 83.24$^{*}$ / -2.21 & 25.36$^{*}$ / 3.44 & 0.83$^{*}$ / 1.68 & 0.22$^{*}$ / 3.71 & 0.87$^{*}$ / 3.87 \\
 &  &  & IQR & (80.94, 88.89) & (25.09, 25.62) & (0.80, 0.84) & (0.19, 0.23) & (0.81, 0.88) \\
 &  & \multirow{2}{*}{\Mf} & median / ES & 79.72$^{*}$ / -2.06 & 25.90$^{*}$ / 2.97 & 0.84$^{*}$ / 1.53 & 0.23$^{*}$ / 3.39 & 0.89$^{*}$ / 3.64 \\
 &  &  & IQR & (78.29, 86.29) & (25.65, 26.35) & (0.81, 0.84) & (0.21, 0.24) & (0.84, 0.90) \\
\midrule
\multirow{12}{*}{\rotatebox{90}{SynthRAD}} & \multirow{6}{*}{\rotatebox{90}{Clinical}} & \multirow{2}{*}{\Ms} & median & 81.55 & \textbf{26.63} & 0.83 & \textbf{0.29} & 0.96 \\
 &  &  & IQR & (66.79, 86.88) & (25.81, 27.97) & (0.81, 0.85) & (0.27, 0.33) & (0.95, 0.97) \\
 &  & \multirow{2}{*}{\Mr} & median / ES & \textbf{77.61$^{*}$} / 0.01 & 26.43$^{*}$ / 0.16 & \textbf{0.84$^{*}$} / -0.02 & 0.27$^{*}$ / 0.95 & 0.96$^{*}$ / 0.09 \\
 &  &  & IQR & (69.61, 82.40) & (25.98, 27.66) & (0.82, 0.84) & (0.26, 0.28) & (0.95, 0.97) \\
 &  & \multirow{2}{*}{\Mf} & median / ES & 80.87$^{*}$ / -0.11 & 26.35$^{*}$ / 0.25 & 0.83$^{*}$ / 0.09 & 0.27$^{*}$ / 0.74 & \textbf{0.96$^{*}$} / -0.07 \\
 &  &  & IQR & (70.68, 87.15) & (25.79, 27.08) & (0.82, 0.85) & (0.26, 0.29) & (0.95, 0.97) \\
\cmidrule{2-9}
 & \multirow{6}{*}{\rotatebox{90}{SynthRAD}} & \multirow{2}{*}{\Ms} & median & 77.09 & 26.83 & 0.84 & \textbf{0.29} & \textbf{0.96} \\
 &  &  & IQR & (61.26, 87.67) & (26.08, 28.40) & (0.82, 0.86) & (0.27, 0.33) & (0.95, 0.97) \\
 &  & \multirow{2}{*}{\Mr} & median / ES & 53.16$^{*}$ / 1.35 & 29.51$^{*}$ / -1.36 & 0.88$^{*}$ / -1.22 & 0.25$^{*}$ / 1.37 & 0.91$^{*}$ / 1.51 \\
 &  &  & IQR & (46.39, 63.05) & (28.19, 30.51) & (0.86, 0.90) & (0.20, 0.27) & (0.82, 0.94) \\
 &  & \multirow{2}{*}{\Mf} & median / ES & 57.00$^{*}$ / 1.19 & 29.48$^{*}$ / -1.34 & 0.87$^{*}$ / -1.10 & 0.25$^{*}$ / 1.50 & 0.90$^{*}$ / 1.53 \\
 &  &  & IQR & (51.86, 68.26) & (27.94, 30.19) & (0.85, 0.89) & (0.20, 0.27) & (0.83, 0.95) \\
  &  & \multirow{2}{*}{\shortstack{challenge \\ winner}} & median& \textbf{49.27} & \textbf{31.02} & \textbf{0.91} & - & - \\
 &  &  & IQR &(40.81, 55.94) & (29.36, 32.47) & (0.87, 0.94) & - & - \\ 
\bottomrule
\end{tabular}
\end{adjustbox}
\end{table}

%% file: discussion.tex
\section{Discussion}
\label{disc}
In our study, a computational efficient framework for simulated \gls{CBCT} generation including heuristic scatter modeling, analytic motion field simulation, and optimized CUDA kernels was successfully implemented. This  enabled practical large-scale dataset generation without the requirement of aligned \gls{CT}-\gls{CBCT} pairs. We demonstrated that the framework is capable to create realistic images which were used to perform a systematic comparison between models trained with real and simulated data.\\
Our results revealed a fundamental tension in supervised \gls{sCT} evaluation. Standard benchmark metrics reward registration bias reproduction: Models trained on real data (\Mr) achieved lower \gls{MAE} precisely because they learned to replicate the spatial errors present in imperfectly registered training pairs. This effect was configuration-dependent. When training and test datasets shared the same registration methodology, networks optimizing for biased ground truth were rewarded, and \Mr achieved superior intensity metrics. However, when registration quality differed between training and testing, this learned bias became misaligned with the test distribution, and \Ms demonstrated advantages in both geometric alignment and intensity metrics. This might occur when externally trained models are deployed at new institutions with significant implications: challenge leaderboards ranking methods by \gls{MAE} or \gls{SSIM} may systematically favor biased models, and reported performance gains may partially reflect better registration bias matching rather than better anatomical reconstruction. Clinical utility, as reflected in our \gls{IQS} results, aligns with geometric fidelity rather than intensity metrics, suggesting current benchmarks inadequately capture clinical value.

The dataset-specific correlation analysis provided direct evidence for this concern. For the clinical dataset employing deformable registration, intensity metrics exhibited inverted correlations with \gls{IQS}. Higher \gls{MAE} was associated with higher clinical ratings ($\rho$ = 0.57), while higher \gls{PSNR} and \gls{SSIM} were associated with lower ratings ($\rho$ = -0.46 and -0.59, respectively). This inversion occurred because deformable registration introduces spatial errors into the ground truth. Models minimizing \gls{MAE} learn to reproduce these errors \cite{Boussot2025}, yielding outputs that score well on benchmarks but were rated poorly by observers. For the SynthRAD dataset with rigid registration, intensity metrics showed no significant correlation with clinical preference, rendering them uninformative rather than misleading. The near-zero correlations observed in the combined analysis result from these opposing dataset-specific effects cancel out, which may obscure metric validity issues in multi-site studies with heterogeneous registration pipelines \cite{Nenoff2023, Rohlfing2012}. In contrast, \gls{NMI} maintained consistent positive correlation with \gls{IQS} across both datasets ($\rho$ = 0.29 and 0.31), supporting its use as a registration-bias-robust evaluation metric.\\

These observations have implications for model evaluation and benchmark design. Current challenges such as SynthRAD \cite{SynthRAD2023} employ consistent registration pipelines across training and validation sets, which may inadvertently reward models that learn registration-specific biases rather than true anatomical transformations. For clinical applications requiring accurate image registration, such as dose accumulation, structure propagation, treatment response assessment or structure segmentation, geometric fidelity may be more important than intensity accuracy alone \cite{Chetty2019, Brion2021, Belfatto2016}. Current state-of-the-art methods reported excellent intensity metrics (HUDiff: MAE 23.31 HU for brain and 26.11 HU for H\&N \cite{Hu2025}; TransCBCT: MAE 28.8 HU for prostate \cite{Chen2022}), but geometric alignment was rarely evaluated, potentially masking clinically significant spatial errors.\\

The quality of ground truth data represents a fundamental, often overlooked factor limiting supervised \gls{sCT} generation. Clinical \gls{CBCT}/\gls{CT} pairs are rarely perfectly aligned due to inter-session variations in patient positioning and anatomy, forcing reliance on registration algorithms that introduce residual spatial errors. Our physics-based \gls{sCBCT} generation framework addresses this limitation by providing perfectly aligned training pairs by construction.\\
Boussot \etal \cite{Boussot2025} observed analogous "registration bias" effects across multiple anatomical sites, finding that models trained with imperfect alignment achieved higher challenge scores precisely because evaluation employed the same registration strategy, directly supporting our findings (see \autoref{fig:nmi_comparison}).\\

Inter-rater reliability for absolute \gls{IQS} was limited (\autoref{tab:rater_stats}), reflecting the subjective nature of image quality assessment and varying emphasis on different quality dimensions among observers \cite{Flight2015}. However, the strong consensus in preference (87\% favoring \Ms, average \gls{IQS} 3.94 versus 2.04) indicated that despite variability in absolute scoring, the relative superiority of geometrically-faithful outputs was clinically perceptible. Observers consistently preferred outputs preserving anatomical relationships over those with lower intensity error but degraded spatial fidelity, suggesting geometric accuracy aligns with clinical intuition in ways intensity metrics alone do not capture. Based on these findings, we recommend incorporating geometric alignment metrics, evaluated against input images rather than registered ground truth, as standard evaluation criteria for \gls{sCT} synthesis. However, geometric alignment metrics have limitations; they will fail to penalize intensity artifacts that do not affect spatial correspondence. Further investigation of evaluation metrics capturing both geometric fidelity and intensity accuracy without dependence on ground truth quality is warranted.\\

Several limitations should be acknowledged. The heuristic scatter model was empirically tuned for Elekta XVI imaging system and may require adjustment for other systems. The sinusoidal respiratory motion model may inadequately represent irregular breathing patterns and was currently also limited to the pelvic region. Clinical validation was limited to a single institution and gynecological cohort; generalization to other vendors and anatomical sites requires further investigation which is planned for follow-up work as well as the support for more body regions such as the abdomen and the thorax. Critically, while we demonstrated improved geometric alignment, downstream dosimetric impact was not directly evaluated which is an essential next step for establishing clinical utility.\\
The framework's ability to generate unlimited geometrically-aligned training data offers practical advantages. Institutions can create site-specific datasets from planning data archives without paired \gls{CBCT} acquisitions. The \Mf results demonstrated that simulated pretraining provided useful initialization refinable with limited real data. Furthermore, our framework extends naturally to segmentation, reconstruction, and registration tasks where ground truth is similarly constrained by registration quality and missing label datasets.\\

%% file: conclusion.tex
\section{Conclusion}

This study demonstrated that physics-based \gls{sCBCT} generation provides a viable solution to the registration problem limiting supervised \gls{sCT} synthesis approaches. By eliminating registration errors through simulation, our framework enabled training with perfect geometric correspondence, resulting in models that preserve spatial relationships more accurately than those trained on registered clinical pairs. The consistent geometric alignment advantages observed across two independent datasets, combined with strong clinical observer preference for simulation-trained model outputs, support the utility of this approach for adaptive radiotherapy applications where spatial accuracy is of outmost importance. The computational efficiency of the framework enabled practical large-scale dataset generation, providing a foundation for further development of geometrically-faithful \gls{sCT} synthesis methods.

%% file: ablation_study.tex
\section{Ablation study}
\setcounter{table}{0}
\label{app1}

We performed an ablation study for adjustable parameters: saturation factor $c_{sat}$, breathing amplitude $A_{max}$, scatter-to-primary ratio $SPR$ and photon flux $\Phi$. Although $\Phi$ and $SPR$ was known for the Elekta XVI system, we evaluated whether the nominal value was optimal for our simulation framework. Each parameter was varied independently while holding others at their default values, evaluated across n=10 patients (\autoref{tab:ablation_detailed}).

  \begin{table}[htbp!]
  \centering
  \caption{Ablation study results with \gls{FID} and \gls{MMD} across n=10 patients.}
  \label{tab:ablation_detailed}
  \begin{tabular}{@{}lcc@{}}
  \toprule
  \textbf{Configuration} & \textbf{FID $\downarrow$} & \textbf{MMD $\downarrow$} \\
  \midrule
  \multicolumn{3}{l}{\textit{Saturation Factor ($\Phi=4.16\times10^5$, $A_{max}=5$ mm, $SPR=1.6$)}} \\
  \quad $c_{sat} = 1.0$ & 69.47 & 2.16 \\
  \quad $c_{sat} = 1.5$ & 68.47 & 1.52 \\
  \quad $c_{sat} = 2.0$ & \textbf{63.62} & \textbf{1.10} \\
  \midrule
  \multicolumn{3}{l}{\textit{Photon Flux ($c_{sat}=2.0$, $A_{max}=5$ mm, $SPR=1.6$)}} \\
  \quad $\Phi = 2.0\times10^5$ & 66.67 & 1.38 \\
  \quad $\Phi = 4.16\times10^5$ & 63.20 & \textbf{1.11} \\
  \quad $\Phi = 6.0\times10^5$ & \textbf{62.66} & \textbf{1.11} \\
  \midrule
  \multicolumn{3}{l}{\textit{Breathing Amplitude ($c_{sat}=2.0$, $\Phi=4.16\times10^5$, $SPR=1.6$)}} \\
  
  \quad $A_{max} = 0$ mm & 71.00 & 1.43 \\
  \quad $A_{max} = 5$ mm & \textbf{63.41} & \textbf{1.12} \\
  \quad $A_{max} = 10$ mm & 70.04 & 1.38 \\
 \midrule
  \multicolumn{3}{l}{\textit{Scatter-to-Primary Ratio ($c_{sat}=2.0$, $\Phi=4.16\times10^5$, $A_{max}=5$ mm)}} \\
  \quad $SPR = 0.5$ & 64.77 & \textbf{1.10} \\
  \quad $SPR = 1.6$ & \textbf{63.57} &	1.11\\
\quad $SPR = 2.5$ &66.03&	1.36\\

  \bottomrule
  \end{tabular}
  \end{table}

Optimal \gls{FID} and \gls{MMD} were achieved with $c_{sat}$ = 2.0 and $A_{max}$ = 5 mm. The nominal photon flux ($\Phi = 4.16 \times 10^5$) performed comparably to higher values, with \gls{MMD} favoring the nominal setting while \gls{FID}/\gls{MMD} showed marginal improvement at $\Phi = 6.0 \times 10^5$ and $SPR=0.5$; we retained the physically measured value. Notably, disabling motion simulation ($A_{max}$ = 0 mm) substantially degraded realism (\gls{FID}: 71.00 vs 63.41), confirming that respiratory motion modeling is necessary for generating realistic simulated \gls{CBCT} images.

%% file: bibliography.bib
@article{DeepDRR2019,
  author       = {Unberath, Mathias and Zaech, Jan-Nico and Gao, Cong and Bier, Bastian and Goldmann, Florian and Lee, Sing Chun and Fotouhi, Javad and Taylor, Russell and Armand, Mehran and Navab, Nassir},
  title        = {{Enabling Machine Learning in X-ray-based Procedures via Realistic Simulation of Image Formation}},
  year         = {2019},
  journal      = {International journal of computer assisted radiology and surgery (IJCARS)},
  publisher    = {Springer},
}

@inproceedings{DeepDRR2018,
  author       = {Unberath, Mathias and Zaech, Jan-Nico and Lee, Sing Chun and Bier, Bastian and Fotouhi, Javad and Armand, Mehran and Navab, Nassir},
  title        = {{DeepDRR--A Catalyst for Machine Learning in Fluoroscopy-guided Procedures}},
  date         = {2018},
  booktitle    = {Proc. Medical Image Computing and Computer Assisted Intervention (MICCAI)},
  publisher    = {Springer},
}

@article{Isensee2021,
author = {Isensee, Fabian and Jaeger, Paul F. and Kohl, Simon A.A. and Petersen, Jens and Maier-Hein, Klaus H.},
issn = {15487105},
journal = {Nature Methods},
number = {2},
pages = {203--211},
pmid = {33288961},
publisher = {Springer US},
title = {{nnU-Net: a self-configuring method for deep learning-based biomedical image segmentation}},
url = {http://dx.doi.org/10.1038/s41592-020-01008-z},
volume = {18},
year = {2021}
}

@Article{Suwanraksa2023,
AUTHOR = {Suwanraksa, Chitchaya and Bridhikitti, Jidapa and Liamsuwan, Thiansin and Chaichulee, Sitthichok},
TITLE = {CBCT-to-CT Translation Using Registration-Based Generative Adversarial Networks in Patients with Head and Neck Cancer},
JOURNAL = {Cancers},
VOLUME = {15},
YEAR = {2023},
NUMBER = {7},
ARTICLE-NUMBER = {2017},
URL = {https://www.mdpi.com/2072-6694/15/7/2017},
PubMedID = {37046678},
ISSN = {2072-6694},
DOI = {10.3390/cancers15072017}
}

@ARTICLE{Hu2024_SynREG,
  
AUTHOR={Hu, Ying  and Cheng, Mengjie  and Wei, Hui  and Liang, Zhiwen },
         
TITLE={A joint learning framework for multisite CBCT-to-CT translation using a hybrid CNN-transformer synthesizer and a registration network},
        
JOURNAL={Frontiers in Oncology},
        
VOLUME={Volume 14 - 2024},

YEAR={2024},

URL={https://www.frontiersin.org/journals/oncology/articles/10.3389/fonc.2024.1440944},

DOI={10.3389/fonc.2024.1440944},

ISSN={2234-943X}}

@article{Thummerer2025,
   author = {Adrian Thummerer and Erik van der Bijl and Arthur Jr Galapon and Florian Kamp and Mark Savenije and Christina Muijs and Shafak Aluwini and Roel J.H.M. Steenbakkers and Stephanie Beuel and Martijn P.W. Intven and Johannes A. Langendijk and Stefan Both and Stefanie Corradini and Viktor Rogowski and Maarten Terpstra and Niklas Wahl and Christopher Kurz and Guillaume Landry and Matteo Maspero},
   doi = {10.1002/mp.17981},
   issn = {24734209},
   issue = {7},
   journal = {Medical Physics},
   month = {7},
   pmid = {40665582},
   publisher = {John Wiley and Sons Ltd},
   title = {SynthRAD2025 Grand Challenge dataset: Generating synthetic CTs for radiotherapy from head to abdomen},
   volume = {52},
   year = {2025}
}

@article{Thummerer2023,
   author = {Adrian Thummerer and Erik van der Bijl and Arthur Galapon and Joost J.C. Verhoeff and Johannes A. Langendijk and Stefan Both and Cornelis (Nico) A.T. van den Berg and Matteo Maspero},
   doi = {10.1002/mp.16529},
   issn = {24734209},
   issue = {7},
   journal = {Medical Physics},
   month = {7},
   pages = {4664-4674},
   pmid = {37283211},
   publisher = {John Wiley and Sons Ltd},
   title = {SynthRAD2023 Grand Challenge dataset: Generating synthetic CT for radiotherapy},
   volume = {50},
   year = {2023}
}

@article{SynthRAD2023,
title = {Generating synthetic computed tomography for radiotherapy: SynthRAD2023 challenge report},
journal = {Medical Image Analysis},
volume = {97},
pages = {103276},
year = {2024},
issn = {1361-8415},
doi = {https://doi.org/10.1016/j.media.2024.103276},
url = {https://www.sciencedirect.com/science/article/pii/S1361841524002019},
author = {Evi M.C. Huijben and Maarten L. Terpstra and Arthur Jr. Galapon and Suraj Pai and Adrian Thummerer and Peter Koopmans and Manya Afonso and Maureen {van Eijnatten} and Oliver Gurney-Champion and Zeli Chen and Yiwen Zhang and Kaiyi Zheng and Chuanpu Li and Haowen Pang and Chuyang Ye and Runqi Wang and Tao Song and Fuxin Fan and Jingna Qiu and Yixing Huang and Juhyung Ha and Jong {Sung Park} and Alexandra Alain-Beaudoin and Silvain Bériault and Pengxin Yu and Hongbin Guo and Zhanyao Huang and Gengwan Li and Xueru Zhang and Yubo Fan and Han Liu and Bowen Xin and Aaron Nicolson and Lujia Zhong and Zhiwei Deng and Gustav Müller-Franzes and Firas Khader and Xia Li and Ye Zhang and Cédric Hémon and Valentin Boussot and Zhihao Zhang and Long Wang and Lu Bai and Shaobin Wang and Derk Mus and Bram Kooiman and Chelsea A.H. Sargeant and Edward G.A. Henderson and Satoshi Kondo and Satoshi Kasai and Reza Karimzadeh and Bulat Ibragimov and Thomas Helfer and Jessica Dafflon and Zijie Chen and Enpei Wang and Zoltan Perko and Matteo Maspero}
}

@article{Rit2014,
doi = {10.1088/1742-6596/489/1/012079},
url = {https://doi.org/10.1088/1742-6596/489/1/012079},
year = {2014},
month = {mar},
publisher = {},
volume = {489},
number = {1},
pages = {012079},
author = {Rit, S and Vila Oliva, M and Brousmiche, S and Labarbe, R and Sarrut, D and Sharp, G C},
title = {The Reconstruction Toolkit (RTK), an open-source cone-beam CT reconstruction toolkit based on the Insight Toolkit (ITK)},
journal = {Journal of Physics: Conference Series}
}

@misc{Boussot2025,
      title={Why Registration Quality Matters: Enhancing sCT Synthesis with IMPACT-Based Registration}, 
      author={Valentin Boussot and Cédric Hémon and Jean-Claude Nunes and Jean-Louis Dillenseger},
      year={2025},
      eprint={2510.21358},
      archivePrefix={arXiv},
      primaryClass={cs.CV},
      url={https://arxiv.org/abs/2510.21358}, 
}

@InProceedings{Pang2024,
author="Pang, Yunkui
and Liu, Yilin
and Chen, Xu
and Yap, Pew-Thian
and Lian, Jun",
editor="Linguraru, Marius George
and Dou, Qi
and Feragen, Aasa
and Giannarou, Stamatia
and Glocker, Ben
and Lekadir, Karim
and Schnabel, Julia A.",
title="SinoSynth: A Physics-Based Domain Randomization Approach for Generalizable CBCT Image Enhancement",
booktitle="Medical Image Computing and Computer Assisted Intervention -- MICCAI 2024",
year="2024",
publisher="Springer Nature Switzerland",
address="Cham",
pages="646--656",
isbn="978-3-031-72104-5"
}

@article{Liu2020,
  author = {Liu, Y. and Lei, Y. and Wang, T. and Fu, Y. and Tang, X. and Curran, W. J. and Liu, T. and Patel, P. and Yang, X.},
  year = {2020},
  title = {{CBCT}-based synthetic {CT} generation using deep-attention cycle{GAN} for pancreatic adaptive radiotherapy},
  journal = {Medical Physics},
  volume = {47},
  number = {6},
  pages = {2472--2483},
  doi = {10.1002/mp.14121},
  pmid = {32141618},
  pmcid = {PMC7762616}
}

@article{Liang2019,
doi = {10.1088/1361-6560/ab22f9},
url = {https://doi.org/10.1088/1361-6560/ab22f9},
year = {2019},
month = {jun},
publisher = {IOP Publishing},
volume = {64},
number = {12},
pages = {125002},
author = {Liang, Xiao and Chen, Liyuan and Nguyen, Dan and Zhou, Zhiguo and Gu, Xuejun and Yang, Ming and Wang, Jing and Jiang, Steve},
title = {Generating synthesized computed tomography (CT) from cone-beam computed tomography (CBCT) using CycleGAN for adaptive radiation therapy},
journal = {Physics in Medicine \& Biology}
}

@misc{Boussot2025a,
      title={IMPACT: A Generic Semantic Loss for Multimodal Medical Image Registration}, 
      author={Valentin Boussot and Cédric Hémon and Jean-Claude Nunes and Jason Dowling and Simon Rouzé and Caroline Lafond and Anaïs Barateau and Jean-Louis Dillenseger},
      year={2025},
      eprint={2503.24121},
      archivePrefix={arXiv},
      primaryClass={cs.CV},
      url={https://arxiv.org/abs/2503.24121}, 
}

@article{Klein2010,
  title={elastix: a toolbox for intensity based medical image registration},
  author={Klein, Stefan and Staring, Marius and Murphy, Keelin and Viergever, Max A and Pluim, Josien PW},
  journal={IEEE Transactions on Medical Imaging},
  volume={29},
  number={1},
  pages={196--205},
  month={January},
  year={2010},
  doi={10.1109/TMI.2009.2035616}
}

@article{Chen2022,
  author = {Chen, X. and Liu, Y. and Yang, B. and Zhu, J. and Yuan, S. and Xie, X. and Liu, Y. and Dai, J. and Men, K.},
  title = {A more effective {CT} synthesizer using transformers for cone-beam {CT}-guided adaptive radiotherapy},
  journal = {Frontiers in Oncology},
  volume = {12},
  pages = {988800},
  year = {2022},
  month = {August},
  day = {25},
  doi = {10.3389/fonc.2022.988800},
  pmid = {36091131},
  pmcid = {PMC9454309}
}

@article{Hu2025,
  author = {Hu, Can and Cao, Ning and Li, Xiuhan and He, Yang and Zhou, Han},
  title = {{CBCT}-to-{CT} synthesis using a hybrid {U}-{N}et diffusion model based on transformers and information bottleneck theory},
  journal = {Scientific Reports},
  volume = {15},
  number = {1},
  pages = {10816},
  year = {2025},
  month = {March},
  day = {28},
  doi = {10.1038/s41598-025-92094-6},
  issn = {2045-2322},
  url = {https://doi.org/10.1038/s41598-025-92094-6}
}

@misc{Rabe2025,
   author = {Moritz Rabe and Christopher Kurz and Adrian Thummerer and Guillaume Landry},
   doi = {10.1007/s00066-024-02277-9},
   issn = {1439099X},
   issue = {3},
   journal = {Strahlentherapie und Onkologie},
   month = {3},
   pages = {283-297},
   pmid = {39138806},
   publisher = {Springer Science and Business Media Deutschland GmbH},
   title = {Artificial intelligence for treatment delivery: image-guided radiotherapy},
   volume = {201},
   year = {2025},
}

@article{DAYARATHNA2024,
title = {Deep learning based synthesis of MRI, CT and PET: Review and analysis},
journal = {Medical Image Analysis},
volume = {92},
pages = {103046},
year = {2024},
issn = {1361-8415},
doi = {https://doi.org/10.1016/j.media.2023.103046},
url = {https://www.sciencedirect.com/science/article/pii/S1361841523003067},
author = {Sanuwani Dayarathna and Kh Tohidul Islam and Sergio Uribe and Guang Yang and Munawar Hayat and Zhaolin Chen}
}

@article{DonaLemus2024,
  author = {Dona Lemus, Olga Maria and Cao, Minsong and Cai, Bin and Cummings, Michael and Zheng, Dandan},
  title = {Adaptive Radiotherapy: Next-Generation Radiotherapy},
  journal = {Cancers},
  volume = {16},
  number = {6},
  pages = {1206},
  year = {2024},
  month = {March},
  day = {19},
  doi = {10.3390/cancers16061206},
  pmid = {38539540},
  pmc = {PMC10968833},
  issn = {2072-6694},
  url = {https://doi.org/10.3390/cancers16061206},
  publisher = {MDPI}
}

@article{Schulze2011,
  author = {Schulze, R and Heil, U and Gross, D and Bruellmann, DD and Dranischnikow, E and Schwanecke, U and Schoemer, E},
  title = {Artefacts in CBCT: a review},
  journal = {Dentomaxillofacial Radiology},
  volume = {40},
  number = {5},
  pages = {265--273},
  year = {2011},
  month = {July},
  doi = {10.1259/dmfr/30642039},
  pmid = {21697151},
  pmc = {PMC3520262},
  issn = {0250-832X},
  url = {https://doi.org/10.1259/dmfr/30642039},
  publisher = {Sage Publications}
}

@article{Nenoff2023,
  author = {Nenoff, Lena and Amstutz, Florian and Murr, Martina and Archibald-Heeren, Ben and Fusella, Marco and Hussein, Mohammad and Lechner, Wolfgang and Zhang, Ye and Sharp, Greg and {Vasquez Osorio}, Eliana},
  title = {Review and recommendations on deformable image registration uncertainties for radiotherapy applications},
  journal = {Physics in Medicine and Biology},
  year = {2023},
  volume = {68},
  number = {24},
  pages = {24TR01},
  doi = {10.1088/1361-6560/ad0d8a},
  month = {Dec},
  day = {13},
  pmid = {37972540}
}

@article{Rohlfing2012,
  author = {Rohlfing, Torsten},
  title = {Image similarity and tissue overlaps as surrogates for image registration accuracy: widely used but unreliable},
  journal = {IEEE Transactions on Medical Imaging},
  year = {2012},
  volume = {31},
  number = {2},
  pages = {153--163},
  doi = {10.1109/TMI.2011.2163944},
  month = {Feb},
  pmid = {21827972}
}

@article{Chetty2019,
  author = {Chetty, Indrin J and Rosu-Bubulac, Mihaela},
  title = {Deformable Registration for Dose Accumulation},
  journal = {Seminars in Radiation Oncology},
  year = {2019},
  volume = {29},
  number = {3},
  pages = {198--208},
  doi = {10.1016/j.semradonc.2019.02.002},
  month = {Jul},
  pmid = {31027637}
}

@article{Brion2021,
title = {Domain adversarial networks and intensity-based data augmentation for male pelvic organ segmentation in cone beam CT},
journal = {Computers in Biology and Medicine},
volume = {131},
pages = {104269},
year = {2021},
issn = {0010-4825},
doi = {https://doi.org/10.1016/j.compbiomed.2021.104269},
url = {https://www.sciencedirect.com/science/article/pii/S0010482521000639},
author = {Eliott Brion and Jean Léger and A.M. Barragán-Montero and Nicolas Meert and John A. Lee and Benoit Macq}
}

@article{Belfatto2016,
  author = {Belfatto, A and Riboldi, M and Ciardo, D and Cecconi, A and Lazzari, R and Jereczek-Fossa, BA and Orecchia, R and Baroni, G and Cerveri, P},
  title = {Adaptive Mathematical Model of Tumor Response to Radiotherapy Based on CBCT Data},
  journal = {IEEE Journal of Biomedical and Health Informatics},
  year = {2016},
  volume = {20},
  number = {3},
  pages = {802--809},
  doi = {10.1109/JBHI.2015.2453437},
  month = {May},
  pmid = {26173223}
}

@article{Flight2015,
  author = {Flight, Laura and Julious, Steven A},
  title = {The disagreeable behaviour of the kappa statistic},
  journal = {Pharmaceutical Statistics},
  year = {2015},
  volume = {14},
  number = {1},
  pages = {74--78},
  doi = {10.1002/pst.1659},
  month = {Jan-Feb},
  pmid = {25470361}
}
